%% file: egpaper_for_review.tex
\documentclass[10pt,twocolumn,letterpaper]{article}

\usepackage{cvpr}
\usepackage{times}
\usepackage{epsfig}
\usepackage{graphicx}
\usepackage{amsmath}
\usepackage{amssymb}
\usepackage{xcolor}
\usepackage{bbm}
\usepackage{multirow}
\usepackage{subcaption}
\usepackage{authblk}

\makeatletter
\renewcommand\AB@affilsepx{ \ \ \ \ \ \ \ \protect\Affilfont}
\makeatother

\usepackage[pagebackref=true,breaklinks=true,letterpaper=true,colorlinks,bookmarks=false]{hyperref}

\cvprfinalcopy 

\def\cvprPaperID{9847} 

\ifcvprfinal\fi
\begin{document}

\title{Identity Preserve Transform: Understand What Activity Classification Models Have Learnt}

\author[1]{Jialing Lyu}
\author[2]{Weichao Qiu}
\author[3]{Xinyue Wei}
\author[2]{Yi Zhang}
\author[2]{Alan Yuille}
\author[1]{Zheng-Jun Zha}
\affil[1]{University of Science and Technology of China}
\affil[2]{Johns Hopkins University}
\affil[3]{Tongji University}
\affil[ ]{
\tt\small sirine@mail.ustc.edu.cn,\quad{\tt\small \{qiuwch, sarahwei0210,edwardz.amg,alan.l.yuille\}@gmail.com}, \quad{\tt\small zhazj@ustc.edu.cn}
}

\maketitle

\begin{abstract}


Activity classification has observed great success recently. The performance on small dataset is almost saturated and people are moving towards larger datasets. What leads to the performance gain on the model and what the model has learnt? In this paper we propose identity preserve transform (IPT) to study this problem. IPT manipulates the nuisance factors (background, viewpoint, etc.) of the data while keeping those factors related to the task (human motion) unchanged. To our surprise, we found popular models are using highly correlated information (background, object) to achieve high classification accuracy, rather than using the essential information (human motion). This can explain why an activity classification model usually fails to generalize to datasets it is not trained on. We implement IPT in two forms, i.e. image-space transform and 3D transform,  using synthetic images. The tool will be made open-source to help study model and dataset design. 
  
\end{abstract}


\input{1_intro.tex}

\input{2_related.tex}

\input{3_method.tex}

\input{4_exp.tex}

\input{5_conclusion.tex}

{\small
\bibliographystyle{ieee_fullname}
\bibliography{egbib}
}

\end{document}

%% file: 1_intro.tex
\section{Introduction}


We have witnessed the rapid development of video activity classification models thanks to the thriving of Convolution Neural Networks in recent years~\cite{wang2016temporal}\cite{carreira2017quo}\cite{zhou2018temporal}. However, a video activity classification model trained on one dataset often failed to generalize to another~\cite{monfort2019moments}. Activity classification model can solve the task easily by fitting to correlated factors. For example, the model can associate the activity with typical backgrounds where the activity happen or clothes style, rather than trying to parse the human pose. This poses a challenge for the vision model, which requires the model to successfully extract abstract information, rather than using low-level features. Collecting a large number of video is far more expensive than collecting images. This makes the challenge even worse. 





Human activity, in most cases, can be defined by the human motion and should not be associated with \textbf{nuisance factors}, such as the human appearance and the environment. Capturing the \textbf{essential factors} (human motion) is vital for a robust activity classification model. Psycho-physical experiments show that human can reliably recognize the activity by watching moving dots. We expect a well-performing activity classification model to have similar properties. Researchers created larger datasets\cite{carreira2017quo}\cite{monfort2019moments} and carefully selected the combination of nuisance factors~\cite{borji2016ilab}\cite{lecun2004learning} to study the model robustness. In addition to these efforts, we propose a framework to understand a model.



We propose a method called {\textbf{Identity Preserve Transform (IPT)}} to inspect what an activity classification model has learnt. IPT is a transform which manipulates nuisance factors of an image, while keeping the essential factors the same. This type of transform is inspired by the image generation procedure. It includes two types: image-space transform and 3D transform. Image-space transform is applied on a generated image, while 3D transform affects an image by directly changing the underlying nuisance factors. We use both image-processing techniques and a computer vision model providing prior knowledge (semantic transform) for image-space transform. In order to achieve 3D transform, we implemented a synthetic data generation pipeline. It takes a combination of rendering parameters to render a realistic virtual human. Nuisance factors can be directly manipulated.





Identity Preserve Transform unveils interesting properties of state-of-the-art models. Take Temporal Segment Network (TSN) as an example. The model is sensitive to small perturbation created by image-processing. Note that we did not specifically target the model with adversarial attack~\cite{goodfellow2014generative}. More interestingly, the quantitative result shows the model makes decision mainly on the object or background of the video, rather than using the human pose. This can be further validated with visualization technique~\cite{zhou2016learning}. This explains why the high performance on the trained dataset does not generalize to other datasets and real-world scenarios. Our powerful synthetic data pipeline enables us to further analyze the relationship between the model performance and certain factors. The observation for synthetic data can be easily verified using a small real dataset. The IPT operates only on the input data regardless of the model architecture, so it can easily be adopted to study other models.

Our contribution can be summarized as follows: 1. We propose identity preserve transform, a method to inspect an activity model using data probes and independent of model architecture. 2. We analyze a state-of-art model and showed it does not classify video activities according to human motion. 3. We collect a synthetic activity video dataset and develop a diagnostic toolkit to perform IPT. The source code will be available to help others understand and develop activity classification models.


\begin{figure*}
    \centering
    \includegraphics[width=\linewidth]{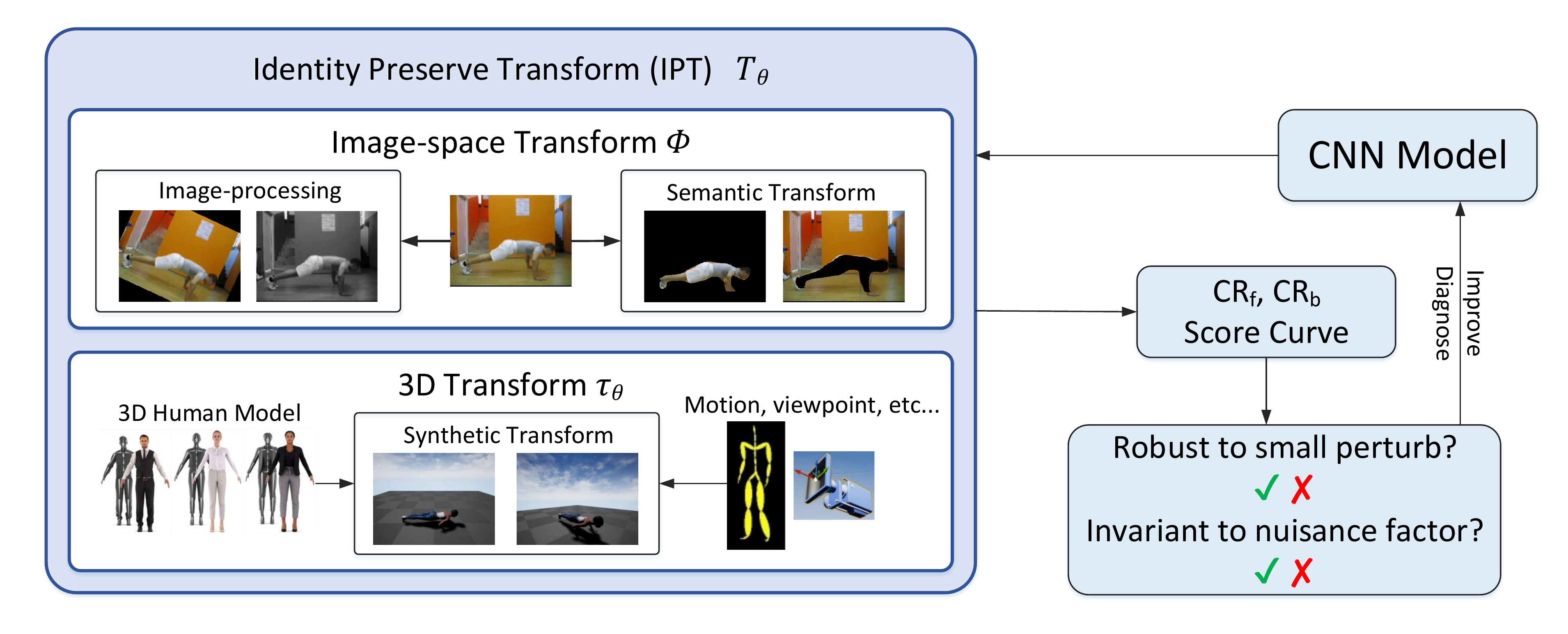}
    \vspace{-0.15in}
    \caption{This figure illustrates an overview of Identity Preserve Transform (IPT). IPT consists of two types. It can take an activity classification model as input and be used to analyze properties of the model. The analysis can be used to improve model design.}
    
    \label{fig:introduce}
\end{figure*}

%% file: 2_related.tex
\section{Related Work}


There are both qualitative and quantitative methods for understanding deep CNNs-based models, the qualitative type mainly focuses on visualizing intermediate layer feature maps and visual saliency, while the quantitative type explains the model through feature importance scores or factor importance scores.


Methods based on feature importance scores alter individual features (pixels, super-pixels, word-vectors, etc) through removal or perturbation~\cite{sundararajan2017axiomatic} for each input to the model to approximate the importance of each feature for model's prediction. They have been proved susceptible to human confirmation biases~\cite{kim2017interpretability}. Whereas Aubry \textit{et al.}~\cite{aubry2015understanding} analyzed CNN feature responses corresponding to different scene factors (object color, object style, lighting, etc.) by controlling them via rendering using a large database of 3D CAD models. In our approach, we abstract influential elements from the generation of human activity videos instead of selecting scene factors.

Most CNN model understanding methods are designed for image inputs. Since our research object is video classification model consuming video inputs, some methods need to be adapted in order to apply while others can not be utilized. On one hand, addition of the time dimension increases the volumn as well as complexity of model input. On the other hand, video classification models are built on multi-branch 2-dimensional CNNs (using 2-D kernels) or 3-dimensional CNNs (using 3-D kernels), making architecture specific methods for image input invalid. Therefore, our main contribution falls on proposing a general method to gain deep understanding into video classification models.

Recently, researchers started to use synthetic data (data generated through computer graphics) to understand vision models. This is mainly due to the high cost and difficulty to collect and annotate large numbers of controlled real data. Synthetic data has been used to study the sensitivity to rendering parameters, such as viewpoint~\cite{zeng2019adversarial}\cite{alcorn2019strike}, material property~\cite{zhang2018unrealstereo}. The controllability of the synthetic data also enables studying the invariance and equivariance property of a model~\cite{aubry2015understanding}. 



%% file: 3_method.tex

\section{Method \label{sec:method}}

\subsection{Identity Preserve Transform}

Our design of Identity Preserve Transform is inspired by the data generation process. There are various factors influencing what the data looks like. Theoretically, all factors can be controlled during the data generation process.

Given a computer vision task, the entire set of factors can be split into 2 subsets. One is those factors directly related to the task which represent the most essential information for the task, such as the human motion for human activity classification in our case, or object shape and texture for object detection. We denote this subset as $p$. The other subset comprises of nuisance factors not directly related to the task, and a model should be insensitive to their change or even to their absence, such as viewpoint, human appearance for human activity classification and background color for object detection. We denote the latter subset as $\theta$. 

Therefore, the data generation process can be modelled by Eq.~\ref{eq:generate}, where $G$ denotes generation function, $I$ denotes the data generated.

\begin{equation}
    I = G(p, \theta)
    \label{eq:generate}
\end{equation}

If we manipulate $p$ with $\tau_p$ and $\theta$ with $\tau_{\theta}$ during data generation process, effects on the generated data $I$ is equivalent to a transform $T$ after the generation, as in Eq.~\ref{eq:allT}.

\begin{equation}
    T(I) = G(\tau_p(p), \tau_{\theta}(\theta))
    \label{eq:allT}
\end{equation}

A well performing model capable of learning the essential information for a specific class (denoted by $p$) should be invariant or insensitive to changes of the nuisance factors (denoted by $\theta$). It means that as long as $p$ keeps constant, no matter how $\tau_{\theta}$ changes $\theta$, a good model should yield same results, as in Eq.~\ref{eq:IPT}. Here we denote the model as $f$.

\begin{equation}
    f(I) = f(T_{\theta}(I)) = f(G(p, \tau_\theta(\theta)))
    \label{eq:IPT}
\end{equation}


Identity Preserve Transform (IPT) refers to transformations that preserves $p$. It can be implemented in the image space or in the 3D space. Hence, IPTs can be categorized into two types: one is image-space transforms, which operate on the generated data $I$, the other is 3D transforms, which operate on $\theta$ during data generation process.

Our goal is to understand video activity classification models. Therefore, we first model the video data generation process to emphasize the identity factor $p$ for activity classification, then introduce approaches for implementing IPT to reach our goal.

\subsection{Modeling Activity Video Generation}

The identity factor $p$ for video activity classification is human motion. It is the most essential information a model should learn from training data, other factors in the video data all belong to nuisance factors $\theta$.

A video expressing a human activity can be produced through the following steps: Position the main person in 3-dimensional space, the person has certain appearance while doing a body motion. Surround the main person with an environment which includes scene background, lighting, and activity related objects. Select a viewpoint that allows seeing the main person. Record an activity video with a camera.

Compared with human motion, background, human appearance, viewpoint and lighting, etc. are all nuisance factors for classifying video human activities. If a human activity classification model can develop a mechanism to model and infer human motion, it should be insensitive to the change in all these nuisance factors.


\subsection{Image-space Transform}

Image-space transform is IPT operating on generated data to simulate the change of nuisance factors during data generation process, as in Eq.~\ref{eq:image-space transform} Image-space means it edits image extracted from a video.

It can be realized by applying image processing techniques commonly used for data augmentation, such as adding noise, blurring, we call it image processing transform. It can also be realized by a computer vision model providing prior semantic knowledge, such as segmenting objects in the image, super resolution, etc., called semantic transform.

\begin{equation}
T_{\theta}(I) = \Phi*G(p, \theta)
\label{eq:image-space transform}
\end{equation}

\subsubsection{Image Processing Transform}

Image processing transform influences how a video look like while keeping background and human motion the same. For example, we can lower the image resolution by applying a blurring filter, or increase the image brightness through histogram equalization. However, lowering the image resolution or increasing brightness would not change people’s body motion inside the video. 

If a video activity classification model is robust enough, we expect its performance not to be harmed by image processing transform on the test data. Once its classifying accuracy drops due to image processing transforms, we should be alarmed that it overfits to image pattern without really learning human motion.

\subsubsection{Semantic Transform}

Semantic transform edits image by an additional computer vision model that provides semantic knowledge while preserving human motion information.

We used Mask-RCNN~\cite{he2017mask} to implement Semantic Transforms in our experiments. As shown in Fig.~\ref{fig:semantic_transform}, Mask-RCNN detects and segments regions of people. With this additional model we can obtain a foreground-only video and a background-only video from the original video by superimposing black masks onto background and foreground respectively. 

We can test a model on the original video set, foreground-only set and background-only set to get three classifying accuracy $Acc_o$, $Acc_f$ and $Acc_b$, then we can compute foreground-only accuracy changing rate and background-only changing rate by Eq.~\ref{eq:changing_rates}.

\begin{equation}
    \begin{cases}
     CR_f=\frac{Acc_o-Acc_f}{Acc_o} \\
                                  \\
     CR_b=\frac{Acc_o-Acc_b}{Acc_o} 
    \end{cases}
    \label{eq:changing_rates}
\end{equation}

$CR_f$ is expected to be very close to 0, if the model has really learnt human motion and use human motion to classify. An ideal segmentation of the human-centric foreground well preserves all kinds of information on the person in foreground-only videos, while leaving a silhouette of the person in background-only videos. Therefore, if the model is capable of infering human motion but has no reliance on background, $CR_b$ should have a positive value closer to 1 rather than 0.

\subsection{3D Transform}
	
A 3D transform is an IPT that directly manipulates nuisance factors during data generation process, as in Eq.~\ref{eq:3D_transform}. It has access to every factor in the data and can manipulate each factor separately.


\begin{equation}
    T_{\theta}(I) = G(p,\tau_{\theta}(\theta))
    \label{eq:3D_transform}
\end{equation}

However, the conditions for achieving 3D-transform in reality are very strict. Previously, researchers use robotic arm in a lab setting to control viewpoint and lighting. Such realization of 3D transform is usually expensive and difficult to set up. Recently, due to the popularity of synthetic data, meaning data generated through computer graphics, researchers started to use synthetic data to control factors of an image, such as viewpoint~\cite{zeng2019adversarial}\cite{alcorn2019strike}, and material property~\cite{zhang2018unrealstereo}.

We collected a 3D animation dataset and created a toolbox for generating synthetic human activity videos. The toolbox called unrealcv can be used to record and render images. After setting human motion with an 3D animation, we could use the toolbox to configure and control each nuisance factor separately, including the environment, lighting, human appearance and viewpoint.

Controlling a certain nuisance factor to change while keeping other factors consistent, we can study how sensitve a model is to this controlled factor， and whether the model’s response follow some kind of rule. For example, viewpoint is determined by camera azimuth angle, camera elevation angle and camera distance. Increase or decrease value of one viewpoint variable gradually while keeping the other two consistent in different videos, then record the model’s classification scores, we can obtain a score curve that shows the model’s response pattern for this factor over a specific activity class. If we repeat the viewpoint variable control experiment on different human appearance, we can see whether the score curve will differ due to human appearance change, leading us to know how the model is influenced by human appearance.

Conclusions drawn from synthetic data can be verified by real data. Synthetic domain and real domain share lots of human motion information that is essential for recognizing human activity. When it is expensive and difficult to collect a large number of control variable videos in real domain , we can instead enlarge the increasing or decreasing step of the controlled variable in different videos, and emphasize on important values that reflect the main trend.

\subsubsection{Synthetic Data Generation}

The synthetic data generation pipeline is built on top of Unreal Engine, a popular game engine for creating 3D video game. We extend the engine to enable nuisance factor control, such as human appearance and pose, in python. Image and ground truth were captured using unrealcv~\cite{qiu2017unrealcv}.

In order to create diverse combination of human appearance and activity, we collected a large synthetic human dataset.
Rigged human models were purchased from Epic Game marketplace. The motion capture data comes from CMU Mocap dataset~\cite{cmumocap}. The human motion sequences were imported and adapted to our 3D human models.

The factor control can be done with python interface. We follow the design of unrealcv. The implemented factor control includes viewpoint, human appearance, human activity. We also implemented domain randomization and ground truth generation to enable training models with our synthetic data.
 

The synthetic data generation and nuisance factor control system can be shared and reproduced easily. This is because the simulator has been packed into a Linux and Windows binary executable, so it can be freely shared without an explicit requirement to repurchase the license for these 3D assets.

%% file: 4_exp.tex
\section{Experiment}

In this section Identity Preserved Transform is applied to the trained Temporal Segment Networks (TSN) introduced in \cite{wang2016temporal} and trained Inflated 3D ConvNet (I3D) introduced in \cite{carreira2017quo}. One on hand, TSN is the representative of state-of-the-art video classification models that uses 2D convolutional kernels in neural network.  On the other hand, I3D is the pioneer in 3D CNNs, giving rise to many adaptations such as S3D~\cite{xie2018rethinking}, I3D-GCN~\cite{wang2018videos}, etc.

We adopted TSN parameters trained on UCF101~\cite{soomro2012ucf101} provided by OpenMMLab and I3D parameters trained on Kinetics~\cite{carreira2017quo} provided by deepmind publicly on Github. Both models’ parameters are pre-trained on ImageNet, input modality to both models is RGB without optical flow.

\subsection{Implementing Image-space Transfrom}

Image-space transform is an Identity Preserve Transform operating on image to simulate nuisance factor change. It can be achieved in the form of image processing and semantic transform.

\subsubsection{Image Processing Transform}

Image processing transform includes image processing techniques widely used for data augmentation. We take 5 common image processing techniques in our experiment. They are applied on UCF101 test set to obtain the top-1 accuracy and top-5 accuracy of TSN classification result on each transformed test set.

Results in Tab.~\ref{tab:image_transform} shows that the model is susceptible to image processing techniques, even if they preserve complete human motion information. Histogram equalization drops the top-1 accuracy by over $10\%$, and adding Gaussian noise drops the top-1 accuracy by about $25\%$. Despite the fact that images transformed by these two techniques still look similar to the original image for human.

\input{fig_table/fig_image_transform.tex}

\input{fig_table/tab_image_transform.tex}

The accuracy drop caused by image processing transform conflicts with the expectation for a robust video activity classification model. The reason might be that TSN trained on UCF101 exhibits high reliance on the image pattern exclusive to UCF101.


\subsubsection{Semantic Transform}

\input{fig_table/fig_semantic_transform.tex}

\begin{figure*}
    \centering
    \includegraphics[width=\textwidth]{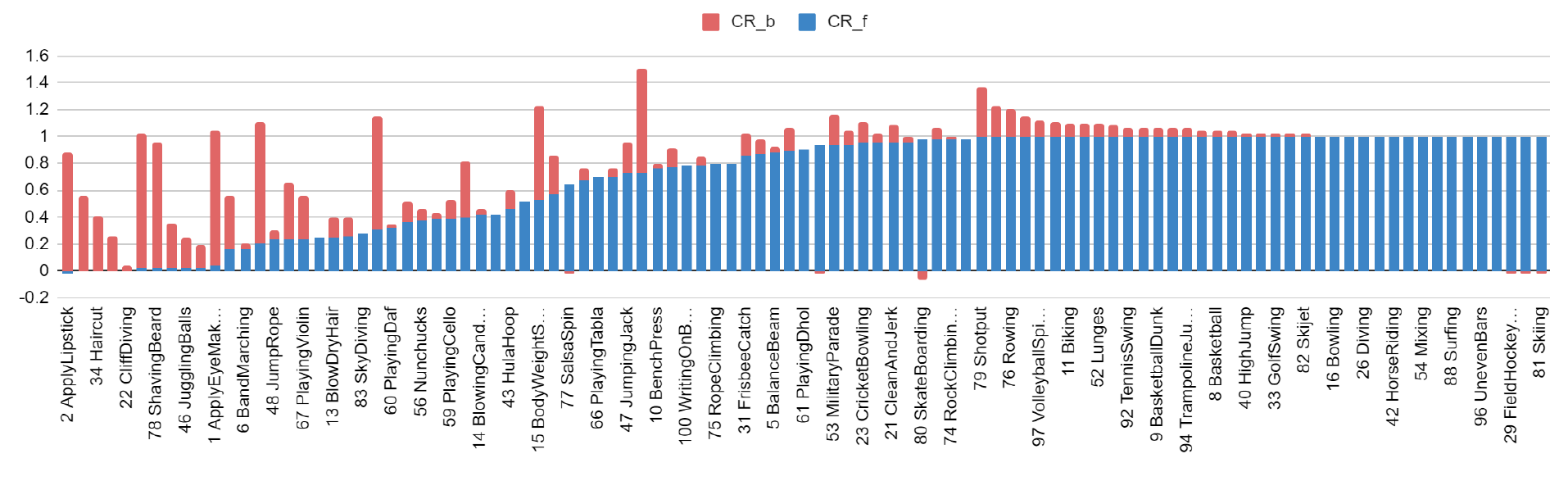}
    \vspace{-0.3in}
    \caption{Plot of $CR_f$ and $CR_b$ for all classes in UCF101, sorted according to the value of $CR_f$. The figure clearly shows that most the model has significant performance drop over most activity classes when only given foreground (high $CR_f$). Only a few activity classes have low $CR_f$, which we further inspected with CAM~\cite{zhou2016learning} visualization.}
    \label{fig:score}
\end{figure*}

Semantic transform edits generated images and videos with a an additional computer vision model that provides prior semantic knowledge. In our experiment we used Mask-RCNN to generate foreground-only videos and background-only videos for UCF101, then tested TSN to get classification accuracy changing rates for each activity class. 

The segmentation outcomes of Mask-RCNN are reasonably good. For rare cases when Mask-RCNN fails to detect human-centric foreground, leading to a full black foreground image and an unmasked background image, we dropped these image pairs. If full black foreground repeatedly appears in one video, we delete video pairs from the test split to ensure the accuracy of our quantitative results.

For a model capable of doing activity classification according to human motion, $CR_f$ is expected to approximate zero while $CR_b$ is expected to have a value closer to 1. However, the real performance is far from perfect. We plotted the $CR_f$ and $CR_b$ given by TSN over all classes of UCF101 and sorted according to the value of $CR_f$ in Fig.~\ref{fig:score}. 
From Fig.~\ref{fig:score} it is easy to tell that the model has significant performance drop over most activity classes when only given foreground (high $CR_f$). Furthermore, there emerges 3 types of results from all UCF101 classes: (1) $1 \approx CR_f \gg CR_b \approx 0$ (2) {\textbf{$0 \approx CR_f \ll CR_b \approx 1$}} (3) otherwise.

Emergence of the first type indicates that for these activity classes, the model has overfit to background rather than learnt human motion. Approximation of $CR_b$ to 0 means the model can classify these activities correctly using only background. Because silhouette of the main person remains in background, if the model successfully recognizes human activities according to the silhouette, it should also achieve high accuracy on foreground-only videos, which better preserves human shape and pose. However, the corresponding $CR_f$ is close to 1, which means accuracy sharply drops on foreground-only videos. Therefore, the model actually does classification using objects and textures in the background.

The quantitative results obtained with semantic transform are supported by class activation mapping~\cite{zhou2016learning}. We generated the class activation maps (CAM) corresponding to the class that gets highest classification score for each video. CAMs of activity classes belonging to type (1) show that the model focuses on objects or textures in the background instead of the human silhouette. For example, for Golf Swing $(CR_f=1,CR_b=0.021)$ the model focuses on the white line on the lawn, and for Archery $(CR_f=0.957,CR_b=0.064)$ the model focuses on the bow, as shown in Fig.~\ref{fig:semantic_transform}.

Simultaneously, the second type of results can be further explained by CAMs. $CR_f$ being close to 0 while $CR_b$ being close to 1 means the model can correctly classify those activity classes with merely foreground, but what specific information is crucial remains unclear. CAMs indicate that the model uses human appearance, objects and textures rather than human motion in the foreground to achieve high accuracy over these classes. For instance, for Apply Eye Makeup $(CR_f=0.041,CR_b=1)$, the model focuses the person's face and the eye brush in the foreground, for Shaving Beard $(CR_f=0.02, CR_b=0.94)$ it focuses on the person’s mouth, as shown in Fig.~\ref{fig:semantic_transform}.

The analysis above points out that TSN trained on UCF101 classify video activities by detecting objects or textures correlated with each activity class in training data. It has not developed a mechanism to model and infer human motion. 

Semantic transform is complementary to class activation mapping. Our approach can be easily scaled up while CAM requires researchers to check a large number of video data with their eyes. Given a large scale video dataset, we can first employ semantic transform to check if the model under examination has overfit to background over some activity classes. After filtering with semantic transform, reviewing work with CAM will significantly decrease.

\subsection{3D Transform}


\input{fig_table/fig_trends.tex}

3D transform is IPT that directly manipulates nuisance factors during data generation process. An ideal 3D transform has access to all factors and can control each one separately, however, it is difficult and expensive to achieve 3D transform on real data. Thus, we first conduct experiments on synthetic data, then verify conclusions drawn from synthetic data with real data.


\begin{figure*}[t]
    \centering
    \includegraphics[width=\linewidth]{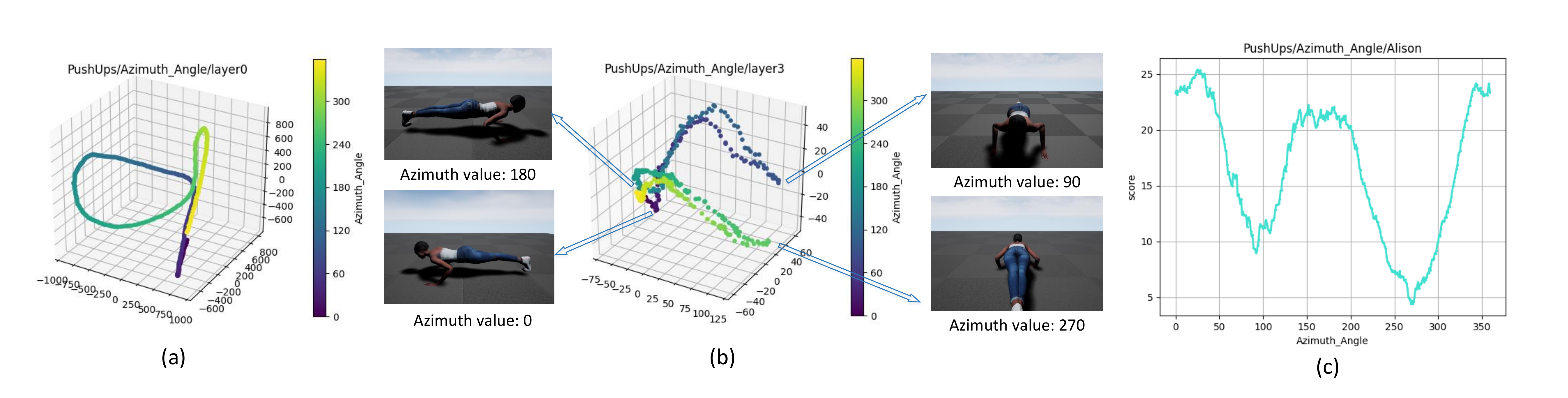}
    \vspace{-0.3in}
    \caption{TSN's response to the azimuth split over Push-ups. TSN is most likely to recognize Push-ups when viewpoint is set towards the flank of person with invariance to left-right flipping, and it can easily fail to recognize Push-ups when viewpoint is set right towards the person's face or feet.(a) PCA embeddings derived from features output by max-pooling layer after the third Inception Module in the Spatial ConvNet of TSN.(b) PCA embedding derived from features output by Segmental Consensus layer. (c) Curve of classification scores output by TSN over the whole azimuth split.
    \label{fig:pca}}
\end{figure*}

\subsubsection{Influence of Viewpoint}

Activity classes for studying viewpoint influence should be defined by human motion that consists symbolic human pose or large body motion, making it possible to be recognized from a large range of viewpoint. From the synthetic video dataset we created with unrealcv toolbox, which has 2548 animation sequences, we chose 6 activities to test I3D and TSN’s classification performance respectively. 

TSN : Push-ups, Archery, Golf Swing, Baseball Pitch Jumping Jack Playing Violin.

I3D : Push-ups, Archery, Climbing Ladder, Dribbling Basketball, Kicking Soccer Ball, Playing Violin

In our synthetic video, the camera is set towards the person, thus viewpoint is determined by three variables: camera azimuth angle, camera elevation angle (altitude angle) and camera distance. Keeping two variable constant while increasing the value of the third one iteratively, we generated a synthetic data split for each viewpoint. Simultaneously, other factors, including human motion, background environment, and human appearance are consistent through out a data split. 

Record a model's classification score corresponding to each viewpoint variable value over a data split, we can obtain a score curve for this variable, as shown in Fig.~\ref{fig:trends}.

If the model is insensitive to viewpoint, the score curve should be stable about a certain score or fluctuate within a relatively small range. However, many score curves yielded by TSN and I3D for different activities are not like this. Instead, there emerged many other trends in these score curves. For example, testing TSN on azimuth split of Jumping Jack videos yields a score curve with two peaks; testing I3D on elevation split of Archery videos yields a score curve monotonically decreasing.

Control variable experiment for multiple activity classes indicate that I3D and TSN are sensitive to viewpoint, the sensitivity extent and trend depends on the controlled viewpoint factor and activity class.

\subsubsection{Influence of Human Appearance}
\input{fig_table/human_appearance.tex}

The model’s response to controlled viewpoint variable can also vary when human appearance changes. We generated the controlled viewpoint variable data splits on different human appearances, then compared the score curves given by the same model over the same activity class. 

Outcomes show that, in some cases the score curve trend on one human appearance no longer appears on another, for example, the score curves obtained from TSN by controlling elevation angle in Push-ups videos look dissimilar on different human appearances, as shown in Fig.~\ref{fig:human_appearance}(b).

However, there also exists cases when the score curves look similar. Take azimuth angle controlled in Push-up videos as example, though human appearance differs, the TSN classification scores are always high when azimuth is around 0 and 180, while being low when azimuth is at 90 or 270, as shown in Fig.~\ref{fig:human_appearance}(a).

Classifying accuracy for different human appearances also differs. We computed the top-1 and top-5 classifying accuracy of TSN and I3D on push-ups videos corresponding to eight different human appearances, as reported in Tab.~\ref{tab:human_appearance}. Both models show preference for some specific human appearances.

\subsubsection{Manifold Visualization Using PCA}

Similar viewpoint score curve corresponding to many synthetic human appearances could probably reflected the model’s classification pattern. Visualizing intermediate layers’ features of the model during a forward pass can help us further understand the outcomes of score curves.

In order to better visualize the manifold which impacts the model performance, we use PCA embedding to visualize the feature of the model. In order to draw more generalized conclusions from azimuth score curves in Push-ups videos shown in Fig.~\ref{fig:pca}, we extracted features from two layers inside the Spatial ConvNet of TSN, then applied dimension reduction using Principal Component Analysis to get low-dimensional features for visualization. The lower layer is max-pooling layer after the third Inception Module~\cite{wang2016temporal}, the higher layer is segmental consensus layer. As Fig.~\ref{fig:pca} shows, 3-dimensional PCA embeddings corresponding to the azimuth angle from 0 to 360 forms a converging manifold, the manifold is twisted through layers between two extraction locations. At Fig.~\ref{fig:pca}(b), embeddings of 0 and 180 azimuth angle have been pulled very close, they represent viewpoint set towards the flank of person, while embeddings of 90 and 270 azimuths have transferred to the other side at 3D feature space, they represent viewpoint set straightly towards the person’s face or feet. 

The visualization indicates high level feature has a certain degree of invariance to viewpoint change.
The feature visualization and score curve together proves that TSN has developed a classification pattern for viewpoint in Push-ups videos. After being trained on UCF101 it has correlated push-ups motion with the human pose looking from a viewpoint at the flank, with invariance to left-right flipping.

\subsubsection{Classification Pattern Supported by Real Data}

\begin{figure}[t]
    \centering
    \includegraphics[width=\linewidth]{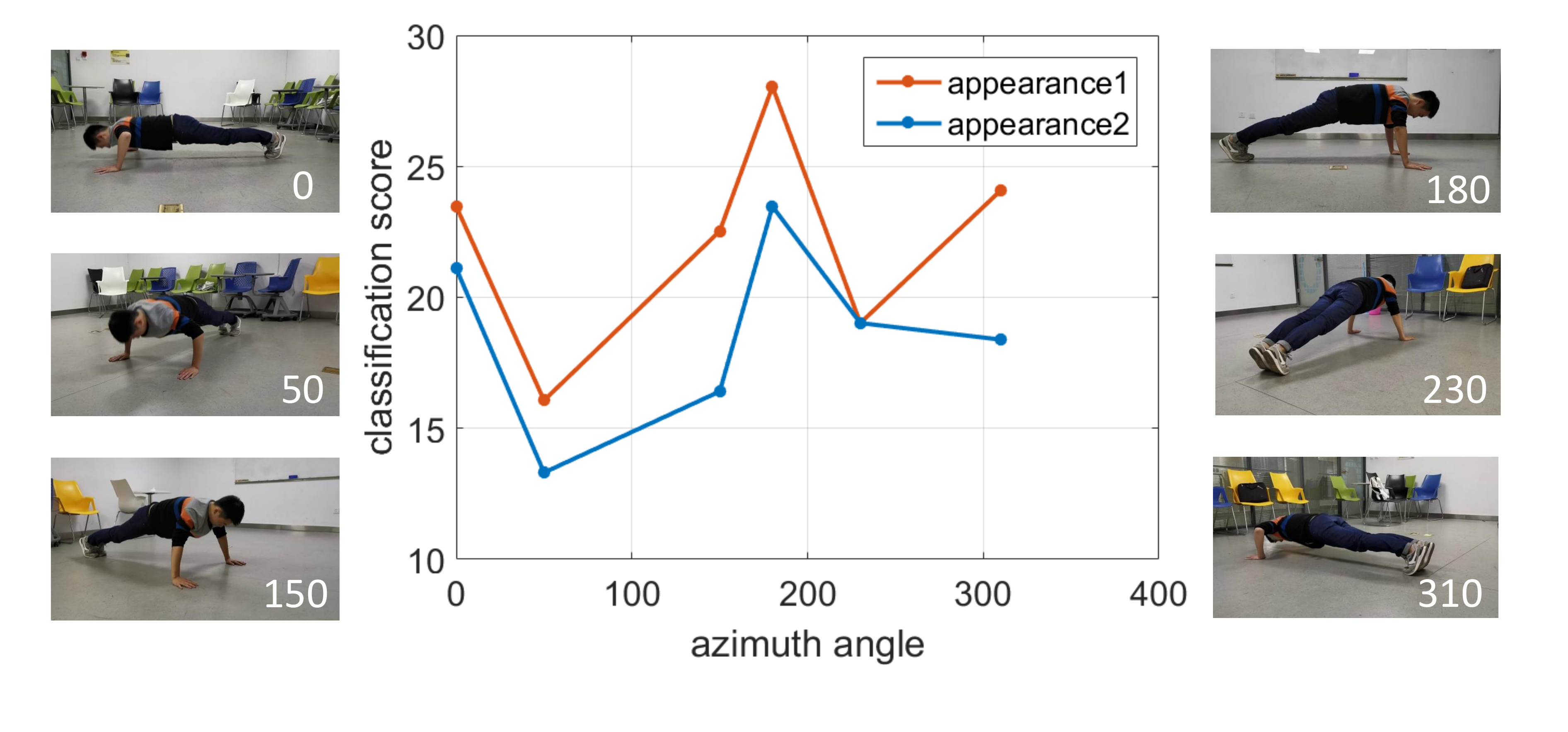}
    \vspace{-0.2in}
    \caption{Classification score curve for real human Push-ups videos follow the same trend emerged from synthetic human Push-ups videos.}
    \label{fig:real_verify}
    
\end{figure}

The classification pattern derived from synthetic data also take effects when the model does classification on real data. Since it is difficult to collect a large number of push-ups videos with different azimuth angles while keeping other factors constant in the real domain, we first recorded push-ups videos from 6 distinct viewpoint at the same height from the same distance, then found out the azimuth angle each real video corresponds to, they are 0, 50, 150, 180, 230, 310 respectively. As expected, the score curve yielded by TSN for these real videos exhibit a trend similar to those over synthetic data. We conducted azimuth controlled experiment on two real human appearances, both come out to be verify the classification pattern found with synthetic data.

%

%% file: fig_table/fig_image_transform.tex
\begin{figure}
    \centering
    \includegraphics[width=\linewidth]{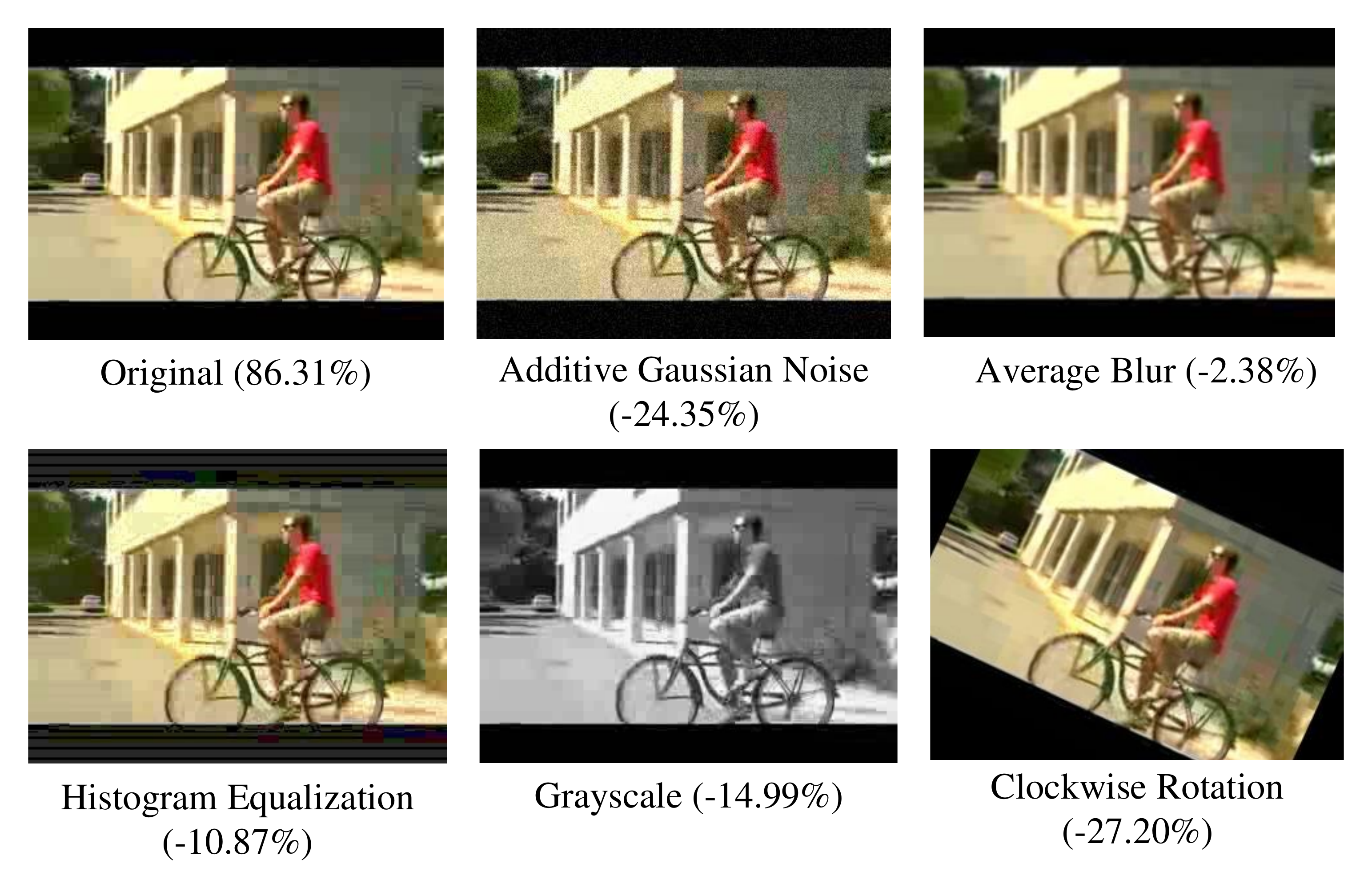}
    \caption{Applying Image-space Transform to videos in UCF101. Image-space transform includes image processing techniques commonly used for data augmentation. Though these image processing techniques preserve human motion information in the video, they lead to serious dropping of TSN's classification accuracy.  \label{fig:image_transform}}
\end{figure}

%% file: fig_table/tab_image_transform.tex
\begin{table}[]
\begin{centering}
\begin{tabular}{|l|l|l|}
\hline
Image-space Transform & Top-1 & Top-5 \\ \hline
Identical Transform               & 86.31\%               & 97.99\%               \\ \hline
Average Blurring                  & 83.93\%               & 96.78\%               \\ \hline
Histogram Equalization            & 75.44\%               & 93.39\%               \\ \hline
Grayscale                         & 71.32\%               & 90.72\%               \\ \hline
Additive Gaussian Noise           & 61.96\%               & 85.41\%              \\ \hline
Clockwise Rotation by 25°         & 59.11\%               & 80.52\%               \\ \hline
\end{tabular}
\caption{Top-1 and top-5 classifying accuracy of trained TSN on a UCF101 test set transformed by 5 different image processing techniques. Different image processing techniques give rise to different extents of decline in classifying accuracy of TSN. Identical transform means the video remains as its original version.  \label{tab:image_transform}}
\end{centering}
\end{table}

%% file: fig_table/fig_semantic_transform.tex
\begin{figure}
    \centering
    \includegraphics[width=\linewidth]{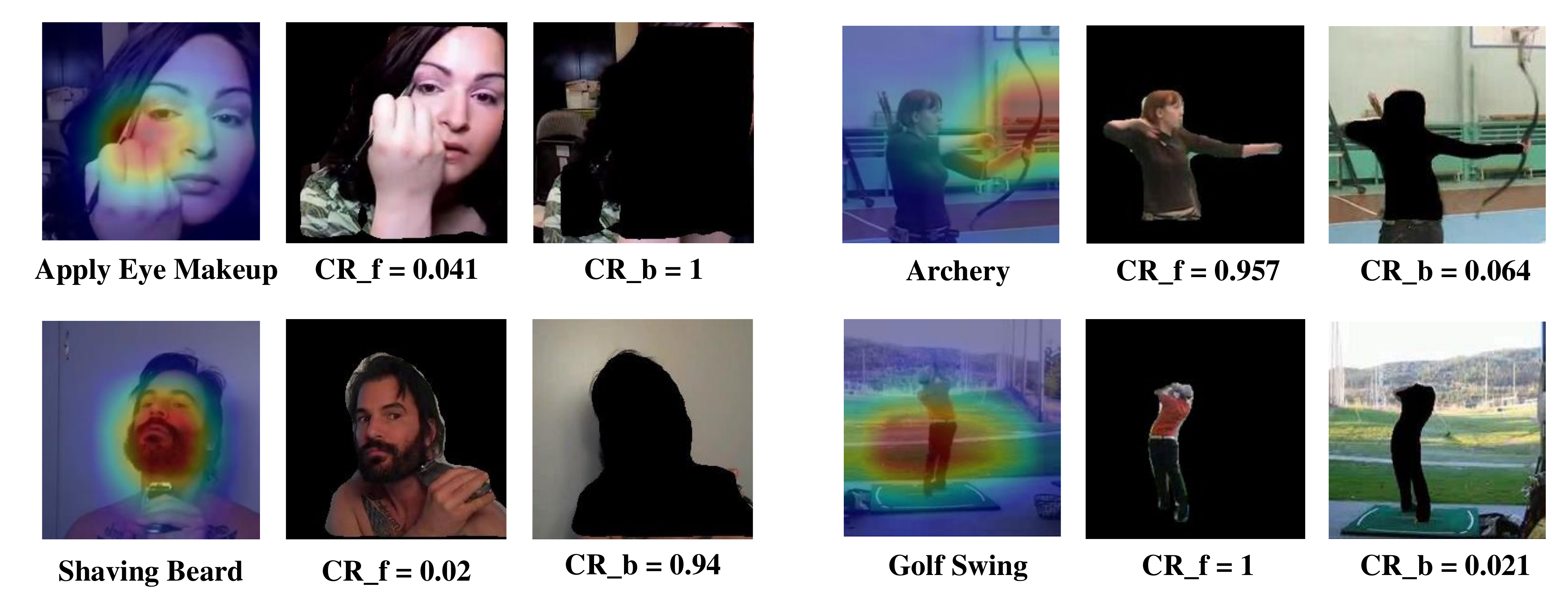}
    \caption{Combined analysis with semantic transform and class activation mapping. Left is the case that $CR_f$ being close to 0 while $CR_b$ being close to 1, the model classify these classes by detecting objects, textures, etc. using human motion information in the foreground. Right is the case that $CR_b$ being close to 0 while $CR_f$ being close to 1, the model overfits to background correlated with the activity classes, these classes can be found easily with semantic transform.
    \label{fig:semantic_transform}
}
\end{figure}

%% file: fig_table/fig_trends.tex
\begin{figure}
    \centering
    \includegraphics[width=\linewidth]{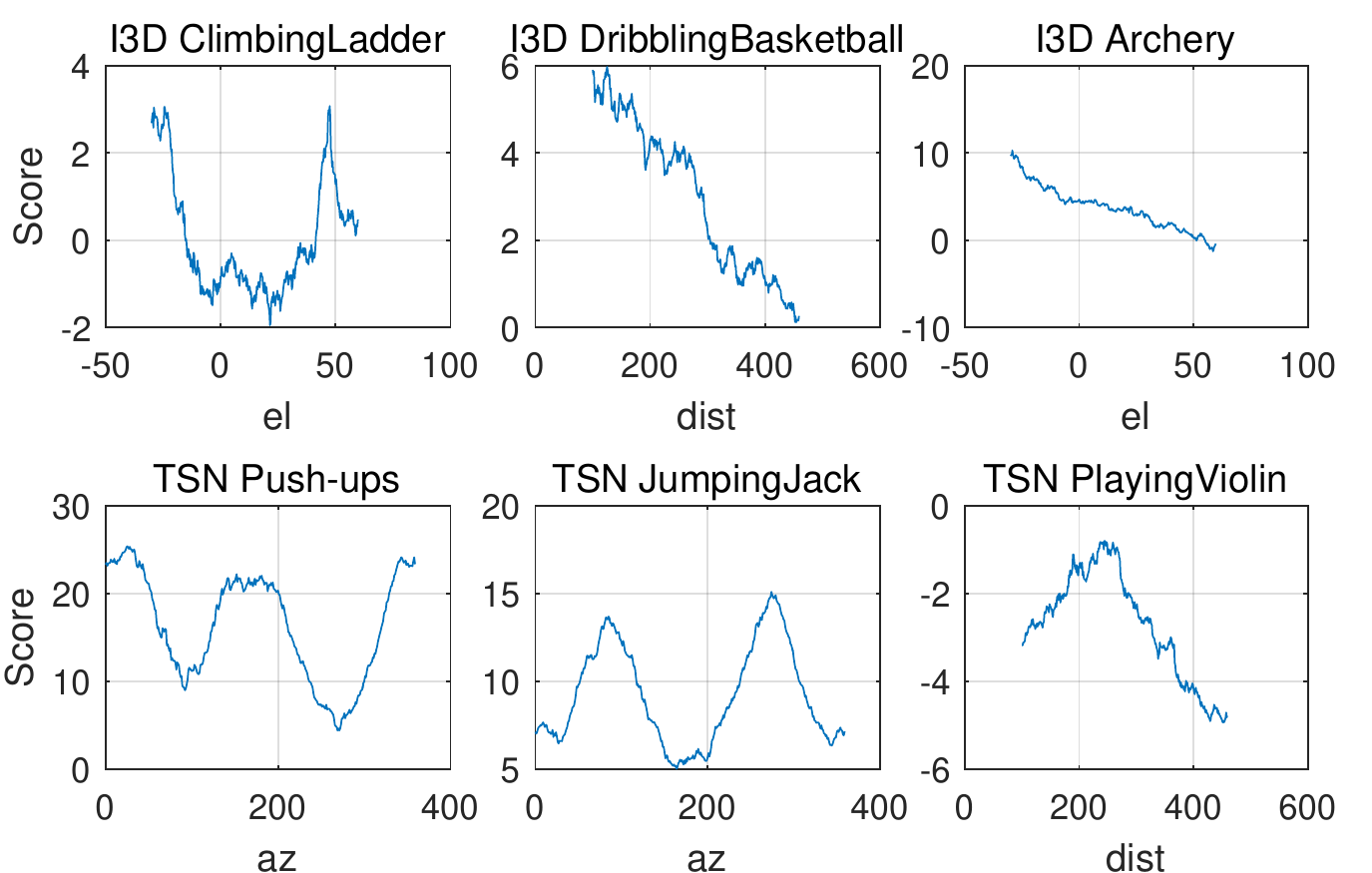}
    \caption{Classification score curves corresponding to different controlled factors obtained by 3D transform. If the model is insensitive to viewpoint, the score curve should be stable about a certain score or fluctuate within a relatively small range. However, many score curves yielded by TSN and I3D display trends other than this. \label{fig:trends}}
\end{figure}

%% file: fig_table/human_appearance.tex
\begin{figure}
    \centering
    \includegraphics[width=\linewidth]{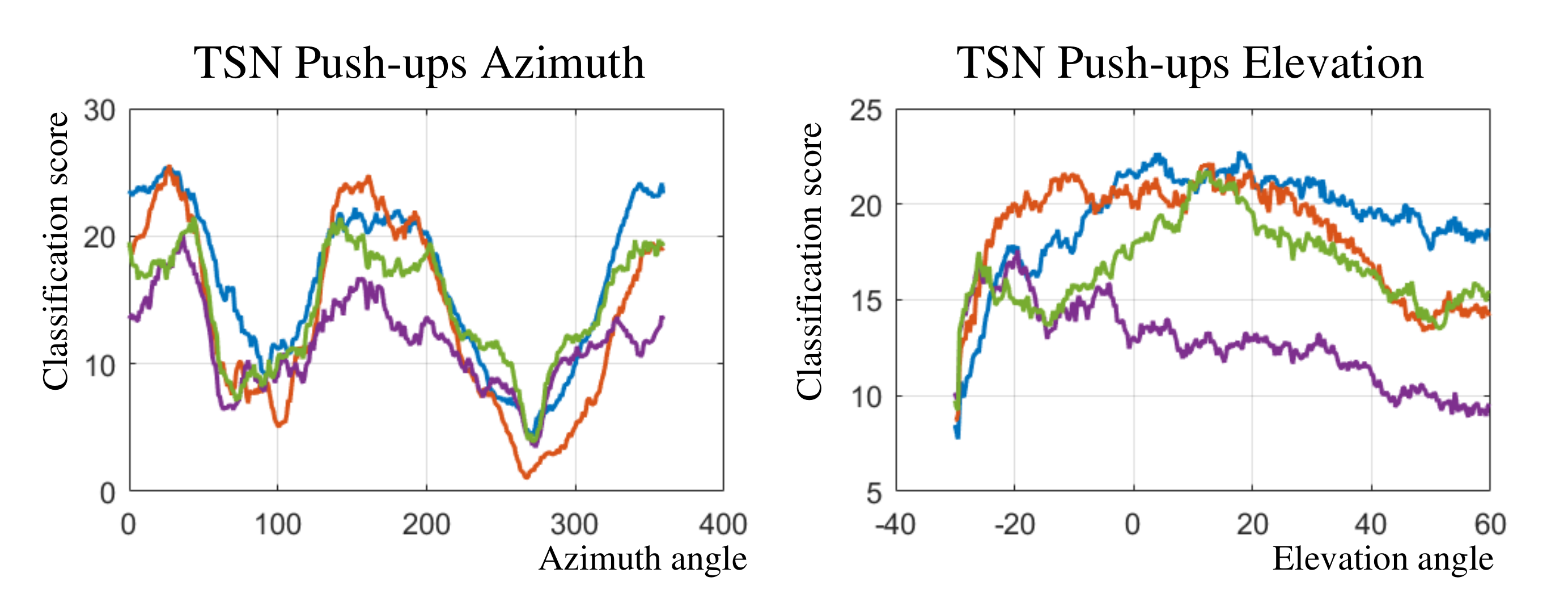}
    \caption{Influence of human appearance on the controlled factor score curve. Each color denotes a synthetic human appearance. In some cases the score curve trend on one human appearance no longer appears on another, yet there also exists cases when the score curves look similar. (a) Though human appearance differs, the TSN classification scores are always high when azimuth is around 0 and 180, while being low when azimuth is at 90 or 270. (b) The score curves obtained from TSN by controlling elevation angle in Push-ups videos look dissimilar on different human appearances
    \label{fig:human_appearance}}
\end{figure}

\begin{table}[]
\begin{tabular}{|c|c|c|c|c|}
\hline
Appearance & Girl\_e & Girl\_f & Girl\_g & Girl\_i \\ \hline
TSN        & 0.19\%  & 0.46\%  & 7.31\%  & 15.09\% \\ \hline
I3D        & 48.61\% & 72.50\% & 70.83\% & 61.39\% \\ \hline
Appearance & Alison  & Carla   & Claudia & Eric    \\ \hline
TSN        & 48.06\% & 39.91\% & 16.20\% & 11.57\% \\ \hline
I3D        & 51.11\% & 61.20\% & 23.24\% & 48.89\% \\ \hline
\end{tabular}
\caption{Top-1 classification accuracy of TSN and I3D on different synthetic human appearances. We aggregated the azimuth, elevation and distance splits generated on the same human appearance into a larger set, yielding 8 human appearance specific sets in total, then tested with TSN and I3D respectively. Difference of classification accuracy between different appearances are obvious, both I3D and TSN show preferences for some appearances.\label{tab:human_appearance}}
\end{table}

%% file: 5_conclusion.tex
\section{Conclusion}

With identity preserve transform we proved that TSN, a state-of-the-art video activity classification model pre-trained on ImageNet and trained on UCF101 is sensitive to factors including the background, related objects, viewpoint and human appearances. Its classification accuracy can be influenced by changes in each of these video factors, therefore TSN probably has memorized the video factor sets for every activity class in UCF101, instead of developing a human motion reasoning mechanism, which is independent of the nuisance factors. Note that our approaches are not only suitable for TSN or I3D. Identity preserve transform is independent of model architecture and thus can be implemented to inspect any trained activity classification model's capabilities, as well as to examine the design of video datasets.